\documentclass{article}
\usepackage{amssymb}
\usepackage{times} 
\usepackage{caption}
\usepackage{float}
\usepackage{amsmath}
\usepackage{array}
\usepackage{appendix}
\usepackage{makecell}
\usepackage{subcaption} 
\usepackage{pifont}
\usepackage{booktabs}
\usepackage{placeins}
\usepackage{multirow}
\usepackage{graphicx}
\usepackage{multicol}
\usepackage{xcolor}
\usepackage{tabularx}

\usepackage[preprint]{corl_2025}

\usepackage[preprint]{corl_2025} 

\title{OmniD: Generalizable Robot Manipulation Policy via Image-Based BEV Representation}

\usepackage{authblk}
\usepackage{bbding} 

\author{
  \textbf{Jilei Mao}$^{1,*}$ \quad
  \textbf{Jiarui Guan}$^{1,*}$ \quad
  \textbf{Yingjuan Tang}$^{1}$ \quad
  \textbf{Qirui Hu}$^{1}$ \quad
  \textbf{Zhihang Li}$^{1}$ \quad
  \textbf{Junjie Yu}$^{1}$ \quad
  \textbf{Yongjie Mao}$^{1}$ \quad
  \textbf{Yunzhe Sun}$^{1}$ \quad
  \textbf{Shuang Liu}$^{1}$ \quad
  \textbf{Xiaozhu Ju}$^{2,\dagger}$ \\
  
   $^1$Beijing Innovation Center of Humanoid Robotics \\
  \texttt{\{lei.mao, julie.tang, qirui.hu, leon.li, luke.yu, jayden.mao, allen.sun, vincent.liu, jason.ju\}@x-humanoid.com,}
  \texttt{jiarui.guan@aalto.fi} 
  
}

\begin{document}
\maketitle

\renewcommand{\thefootnote}{\fnsymbol{footnote}}
\footnotetext[1]{Equal contribution. $^\dagger$~Corresponding author: Xiaozhu Ju (jason.ju@x-humanoid.com).}


\begin{abstract} The visuomotor policy can easily overfit to its training datasets, such as fixed camera positions and backgrounds. This overfitting makes the policy perform well in the in-distribution scenarios but underperform in the out-of-distribution generalization. Additionally, the existing methods also have difficulty fusing multi-view information to generate an effective 3D representation. To tackle these issues, we propose Omni-Vision Diffusion Policy (OmniD), a multi-view fusion framework that synthesizes image observations into a unified bird’s-eye view (BEV) representation. We introduce a deformable attention-based Omni-Feature Generator (OFG) to selectively abstract task-relevant features while suppressing view-specific noise and background distractions. OmniD achieves 11\%, 17\%, and 84\% average improvement over the best baseline model for in-distribution, out-of-distribution, and few-shot experiments, respectively. Training code and simulation benchmark are available on \href{https://github.com/1mather/omnid.git}{GitHub}.
\end{abstract}



\section{Introduction} 

Ensuring robust generalization across diverse environments and scenarios remains a central challenge for real-world embodied systems. The generalization challenges primarily manifest in positional variations, background interference, viewpoint shifts, morphological differences, illumination changes, and environmental dynamics\cite{lin2024data,xie2023decomposing}. To provide a clearer critique for the model's generalization capability, inspired by \cite{kang2024far}, we formally define in-distribution (ID), out-of-distribution (OOD) evaluations, and combinatorial-distribution (CD) for embodied scenarios. 
\begin{figure}[htbp]
    \centering
    \begin{subfigure}[b]{0.3\textwidth}
        \includegraphics[width=\textwidth]{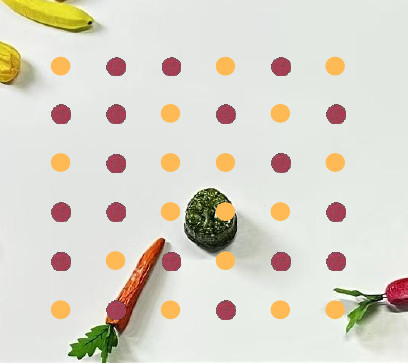}
        \caption{In-distribution}
        \label{fig:subfig1}
    \end{subfigure}
    \hfill
    \begin{subfigure}[b]{0.3\textwidth}
        \includegraphics[width=\textwidth]{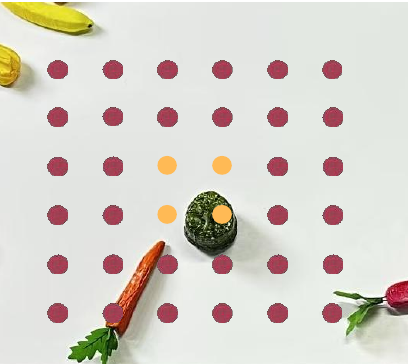}
        \caption{Out-of-distribution}
        \label{fig:subfig2}
    \end{subfigure}
    \hfill
    \begin{subfigure}[b]{0.3\textwidth}
        \includegraphics[width=\textwidth]{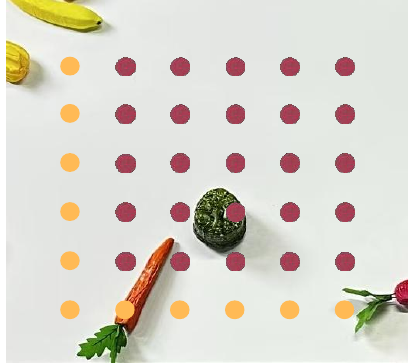}
        \caption{Combinatorial-distribution}
        \label{fig:subfig3}
    \end{subfigure}
    \caption{\textbf{Object position generalization.}The yellow dot denotes the object position in the training data. The red dot denotes the object position in the testing data.}
        
    \label{fig:mainfig}
\end{figure}

Taking object position generalization as an example, as shown in \autoref{fig:mainfig}: when the spatial distribution of pumpkins in test data aligns with the training distribution, it constitutes an ID scenario; significantly divergent distributions indicate OOD cases, while intermediate variations correspond to CD with varying discrepancy levels. Building upon this generalization capability formalization, we systematically evaluate existing methodologies' effectiveness.

Methods like DP\cite{ryu2024diffusion}, ACT\cite{gervet2023act3d}, etc \cite{black20240pi0} are capable of performing complex manipulation tasks and get a high ID success rate. They are prone to overfit to the specific ID scenario and fail to generalize to OOD. Even minor camera pose perturbations or subtle background variations can lead to significant performance degradation.

Methods taking advantage of point clouds~\cite{ze20243d,lu2025manicmrealtime3ddiffusion,ze2025generalizablehumanoidmanipulation3d,cao2024mambapolicyefficient3d,yan2024dnactdiffusionguidedmultitask} achieve partial viewpoint OOD generalization through extrinsic-parameter-driven projection of 3D point clouds into robotic coordinate systems or fixed reference planes. Point clouds filtering enables background OOD generalization by maintaining critical features while suppressing noise, enhancing robustness to environmental variations at the cost of increased computational demands in dynamic environments. While their inherent structural advantages in 3D geometric representations demonstrate enhanced positional OOD generalization compared to 2D approaches, these systems fundamentally depend on high-quality depth data and high-precision calibration. 
\begin{figure*}[htpb]
\centering
\includegraphics[width=135mm]{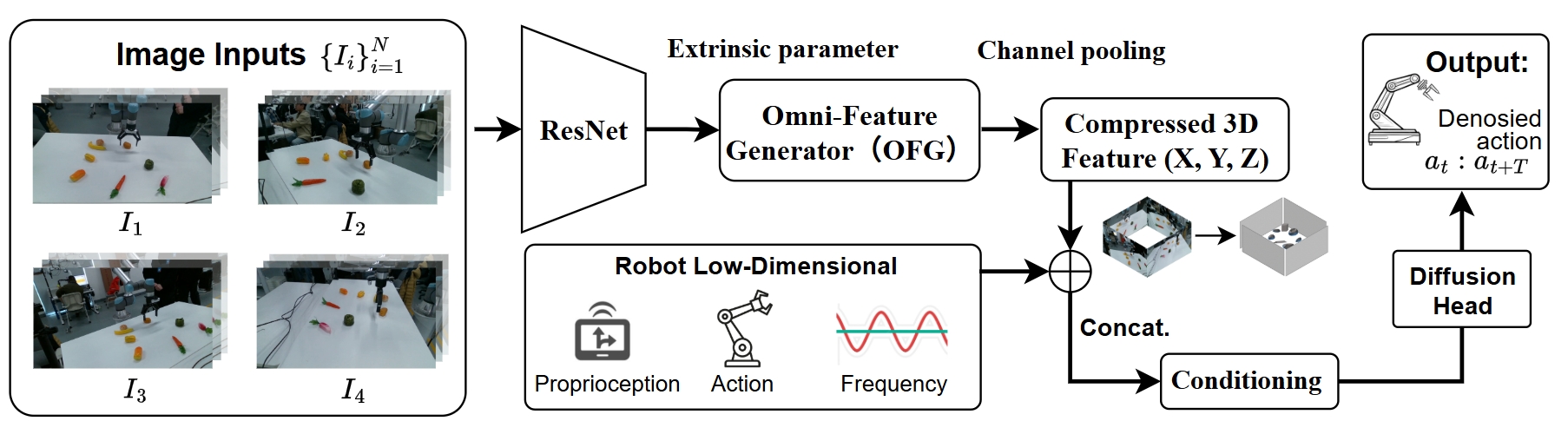} 
\caption{The overall framework of the proposed OmniD method.}
\label{fig01}
\vspace{-5mm} 
\end{figure*}

The rapid expansion of open-source robotic datasets\cite{o2023open-x,khazatsky2024droid} has brought the critical challenge of leveraging these vast data resources to the forefront. Recent methodologies, including HPT\cite{wang2024scalinghpt}, RT-x\cite{brohan2023rt}, Octo\cite{team2024octo}, OpenVLA\cite{kim2024openvla}, $\pi$0 \cite{black20240pi0}, and RDT\cite{liu2024rdt}, have adopted a paradigm of large-scale pretraining followed by task-specific fine-tuning to transfer pre-trained model capabilities to targeted tasks or scenarios. However, reproducing their reported accuracy remains a persistent challenge, and downstream fine-tuning not only demands substantial training resources and specialized optimization techniques, but also yields marginal performance gains in ID and OOD evaluations compared to models like ACT and DP.

In summary, current models face three main challenges: 1) Point clouds methods rely on high-precision depth sensing and precise sensor calibration while suffering from the lack of large-scale datasets. 2) Image-based approaches frequently exhibit severe overfitting issues. 3) Pretraining and fine-tuning paradigms demand substantial training resources yet yield suboptimal performance outcomes. 

In this work, we propose OmniD, a novel multi-camera fusion framework. By introducing the Omni-3D Feature-Generator, OmniD achieves better ID and OOD performance using only monocular RGB images. Furthermore, OmniD exhibits excellent few-shot adaptability, enabling efficient fine-tuning to accommodate novel camera perspectives and tasks. This greatly lowers the computational burden and improves its applicability in real-world scenarios.

The code and models can be found here: \href{https://github.com/1mather/omnid.git}{Omnid}. 
To further support the community's exploration of model generalization, we will also release the Omni Simulation Benchmark, which provides a comprehensive evaluation of positional, background, and viewpoint generalization.

\section{Related work} 
\paragraph{Visuomotor Policy for Robot Manipulation.}
End-to-end visuomotor policies are promising for conducting complex tasks. These policies directly map sensory inputs to actions by imitation learning via expert demonstrations \cite{levine2016endtoendtrainingdeepvisuomotor}. Image-based methods like Diffusion Policy \cite{chi2023diffusion}, ACT \cite{gervet2023act3d},  etc \cite{finn2017oneshotvisualimitationlearning,lee2024behaviorgenerationlatentactions,brohan2023rt,bharadhwaj2024roboagent} have successfully handled complex, long-horizon tasks. However, they lack the generalization to different camera poses and backgrounds since they can easily overfit to specific viewpoint–action pairs \cite{paradis2021intermittent}. The success rate is difficult to guarantee when the camera position or the background changes. 
\vspace{-4mm} 
\paragraph{3D Representations for Robot Manipulation.}
Due to the demands for high precision and generalization in robot manipulation, recent work has focused on 3D scene representations to improve robustness across varying backgrounds and viewpoints~\cite{lu2025manicmrealtime3ddiffusion,ze2025generalizablehumanoidmanipulation3d,cao2024mambapolicyefficient3d,yan2024dnactdiffusionguidedmultitask}. Among these, common representations include RGB-D images~\cite{gervet2023act3d}, point clouds~\cite{ke20243d}, and voxel grids~\cite{shridhar2022perceiveractormultitasktransformerrobotic,james2022coarsetofineqattentionefficientlearning}.

Methods operating on voxel grids, such as PerAct\cite{shridhar2022perceiveractormultitasktransformerrobotic} and C2F-ARM\cite{james2022coarsetofineqattentionefficientlearning}, effectively capture spatial structure but become computationally expensive as resolution increases. Point clouds based approaches, like DP3D \cite{ke20243d}, offer greater efficiency but tend to be sensitive to noise and may lack global scene understanding. Act3D\cite{gervet2023act3d} circumvents voxelization by representing the environment as a continuous 3D feature field, which enhances scalability and expressiveness, albeit at the cost of increased model complexity and training overhead.  In contrast, BEV fusion methods\cite{li2022bevformerlearningbirdseyeviewrepresentation,Huang2022BEVDet4DET}—widely adopted in 3D object detection—demonstrate strong capability in integrating both image-based features and 3D spatial information. Yet, despite their success in perception tasks, no prior work has explored policy learning directly within this BEV feature space.
\vspace{-4mm} 

\paragraph{Foundation Models for Robotics.}
Recent advancements in Vision-Language-Action (VLA) models have revolutionized end-to-end robotic manipulation through unified multimodal foundation model training \cite{qu2025spatialvlaexploringspatialrepresentations,li2024cogactfoundationalvisionlanguageactionmodel,huang2025ottervisionlanguageactionmodeltextaware,liu2025hybridvlacollaborativediffusionautoregression,alayrac2022flamingo,liu2024rdt,team2024octo,black20240pi0,kim2024openvla}. By leveraging pretrained vision-language models (VLMs)\cite{Qwen-VL,Qwen2.5-VL,liu2023llava,liu2023improvedllava,liu2024llavanext,Qwen2-VL} and jointly training on the visual observations, language instructions, and action trajectories, these models achieve unprecedented task generalization in unstructured environments. 

 Despite the progress, the current large VLA models remains constrained by excessive data requirements and prohibitive computational costs for real-world deployment. The reproducibility and downstream fine-tuning performance are also fall below expectations. 
\vspace{-4mm} 
\paragraph{Deformable attention for multi-view fusion.} 
Deformable attention is widely used in autonomous driving to fuse multi-view image features into a BEV representation \cite{sun2024sparsedrive,li2024bevformer,hu2023planningorientedautonomousdriving,lin2022sparse4d,wang2021detr3d}. This representation does not rely on precise LiDAR data and significantly improves performance on downstream tasks such as 3D detection, occupancy prediction, and planning \cite{liang2022bevfusion,hu2023planningorientedautonomousdriving}.

Similar to autonomous driving, robot manipulation tasks accepts multi-view image inputs and requires strong 3D perception. Better 3D representation can improve the module's performance in position and background generalization \cite{gervet2023act3d,ke20243d}. We also find that BEV representation enables efficient fine-tuning to new tasks and viewpoints \cite{liu2023bevfusion}. Based on this, we propose OmniD, a deformable-attention-based multi-view fusion framework. Without relying on high-precision depth or cost-intensive point cloud inputs, OmniD surpasses current state-of-the-art methods in position and background OOD evaluations, while also enabling efficient fine-tuning across diverse tasks and camera viewpoints.
\section{Omni-Vision Diffusion Policy} 

In this section, we will introduce our OmniD framework, dataset, and training configuration. OmniD processes multi-view input through three stages, as shown in \autoref{fig01}: 3D feature learning, dimensional compression, and conditional denoising. All images $\{I_i\}_{i=1}^N$ pass through a shared ResNet-18 backbone to generate 2D features. Then OFG is applied to generate the Omni 3D feature. To enable efficient processing, we compress the 3D representation pooling along the channel dimension. The compressed 3D feature is flattened and concatenated with robot states, forming the conditional input to our diffusion-based action prediction head. The details of OFG and the action head will be introduced in the following section. 
\begin{figure*}[htpb]
\centering
\includegraphics[width=135mm]{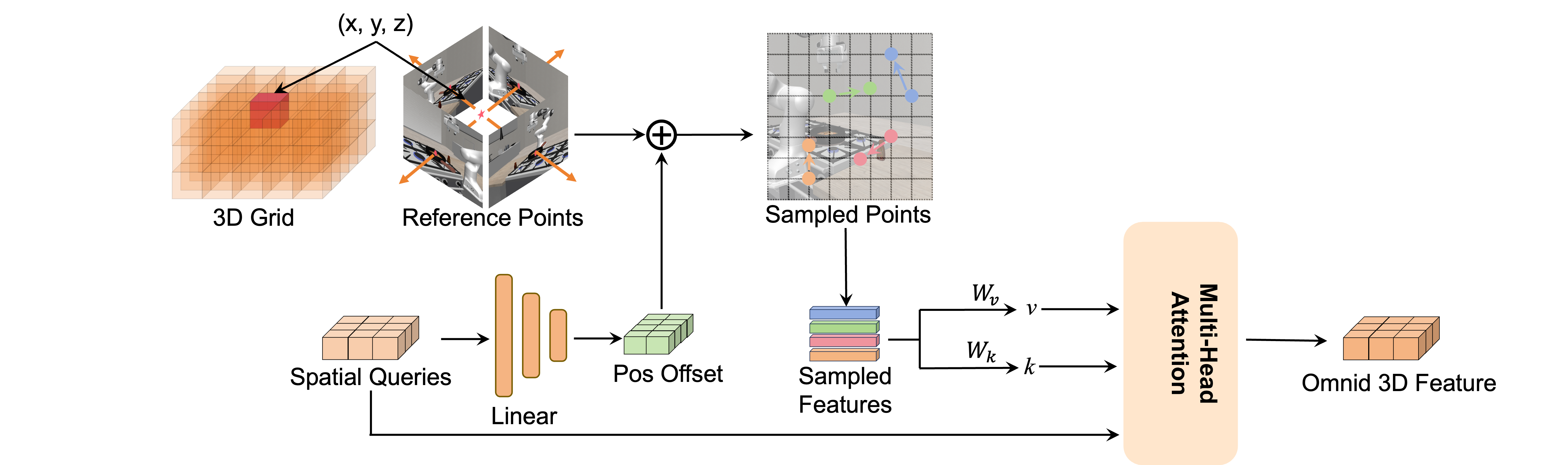} 
\caption{OmniD Feature Generator (OFG).}
\label{fig1}
\vspace{-3mm} 
\end{figure*}
\subsection{Scene Representation with Omni 3D Feature Generator (OFG)}
The proposed OFG is based on \textit{Deformable Attention} and is used to generate 3D spatial features in the BEV space, as illustrated in \autoref{fig1}. The method effectively aggregates image features from multiple camera views by learning spatial queries and their corresponding sampling locations through a deformable attention mechanism.

\subsubsection{Query Generation and Projection}
The BEV range is defined as $(X_{\text{min}}, X_{\text{max}}) = (0,\ 1.152)$, $(Y_{\text{min}}, Y_{\text{max}}) = (-0.64,\ 0.64)$, and $(Z_{\text{min}}, Z_{\text{max}}) = (0,\ 0.768)$, with all values in meters. 
The resolution of the voxel is set to $(\Delta X, \Delta Y, \Delta Z) = (0.018,\ 0.08,\ 0.012)$ meters.
Let $\mathcal{Q} = \{q_i\}_{i=1}^{N}$ denote the set of BEV spatial queries, where $N = 64 \times 16 \times 64$ corresponds to the total number of grid locations in the 3D voxel space and each $q_i \in \mathbb{R}^d$ is a $d$-dimensional embedding corresponding to a reference location $q_i'$. Each reference point $q_i'$ is projected onto the image plane using the camera projection matrices $\{P_j\}_{j=1}^{M}$ from $M$ different views. The projected reference point in image $j$ for query $q_i$ is computed as:
\[
r_{ij} = \pi_j(q_i') = \frac{1}{z_{ij}} K_j [R_j \mid t_j] q_i'
\]
where $K_j$ is the intrinsic matrix, $[R_j \mid t_j]$ is the extrinsic matrix of camera $j$, and $z_{ij}$ is the depth value used for normalization. Here, $q_i$ denotes the spatial query embedding, and $r_{ij}$ represents the 2D projected reference point of $q_i'$ in image $j$.

\subsubsection{Offset Prediction and Sampling}

For each query $q_i$, we use a lightweight network to predict a set of $K$ offsets $\{\Delta p_{ijk}\}_{k=1}^{K}$ for each camera view $j$, generating sample locations:
\[
s_{ijk} = r_{ij} + \Delta p_{ijk}, \quad k = 1, \dots, K
\]
These sampling points are used to extract features from the multi-view images. The offsets $\Delta p_{ijk} \in \mathbb{R}^2$ are learnable and enable adaptive sampling in the image space. The $\Delta p_{ijk}$ is a learnable offset for query $q_i$ in image $j$ and sample index $k$.The $s_{ijk}$ represents the sampling location in image $j$.

\subsubsection{Deformable Attention for Feature Aggregation}

Given the sample locations $\{s_{ijk}\}$ and corresponding image features $\{F_j\}_{j=1}^{M}$, the deformable attention module computes the output feature for each query $q_i$ as:
\[
f_i = \sum_{j=1}^{M} \sum_{k=1}^{K} w_{ijk} \cdot \text{Bilinear}(F_j, s_{ijk})
\]
where $w_{ijk}$ are the attention weights predicted for each sampling location, and $\text{Bilinear}(\cdot)$ denotes bilinear interpolation to sample features from the image feature map. $F_j$ is the feature map of camera image $j$. The $w_{ijk}$ is the attention weight for sample $s_{ijk}$, and $f_i$ is the fused feature for query $q_i$.

This approach enables efficient and flexible aggregation of multi-view image features by leveraging spatial priors in BEV and the adaptability of deformable attention.

\label{OFGM}


\subsection{Policy Representation: Omni 3D Feature in Diffusion policy}
 Leveraging Omni 3D Features, which aggregate multi-view inputs, OmniD models a continuous conditional action distribution. The process initiates from a fully noisy action and iteratively denoises it to yield a clean, executable output:
\begin{equation}
a_t = \sqrt{\bar{\alpha}_t} \, a_0 + \sqrt{1 - \bar{\alpha}_t} \, \epsilon
\end{equation}

where $a_t $ is the noisy action at denoising step $t$. The $a_0$ is the original clean action, $ \epsilon \sim \mathcal{N}(0, I)$ is standard Gaussian noise, and $\bar{\alpha}_t$ is a predefined noise scheduling parameter that controls the noise level at step $k$.
The $g_\theta\left(a_{t}, t\right)$ is trained to predict the noise $\epsilon$ added at each step, minimizing the following objective:
\begin{equation}
\mathcal{L}=\mathbb{E}\left[\left\|\epsilon-g_\theta\left(a_{t}, t\right)\right\|^2\right].
\end{equation}
Here $\mathcal{L}$ is the mean squared error (MSE) loss, $g_\theta$ is the parameterized denoising network of $\theta$. The $a_{t}$ is the noisy input, $t$ is the corresponding timestep, and $\epsilon$ is the ground-truth noise.

\subsection{Omini Multi-view dataset}
Our Omni Multi-view dataset comprises 3000 simulated expert episodes and 20,481 real robot expert trajectories. Every trajectory includes five different cameras. The duration of each task ranges from 5 to 15 seconds. 


For simulation data, we collected 3,000 expert episodes across six tasks: \textit{Select Fruit}, \textit{Add Condiment}, \textit{Select Chemistry Tube}, \textit{Get Coffee}, \textit{Set Study Table}, and \textit{Texas Holdem}, within the customized VLABench~\cite{zhang2024vlabenchlargescalebenchmarklanguageconditioned} simulation environment. These tasks span a diverse set of challenges, including 3D object manipulation (\textit{Add Condiment}, \textit{Select Chemistry Tube}), short-horizon interaction (\textit{Select Fruit}), multi-object scene (\textit{Texas Holdem}), and long-horizon planning (\textit{Get Coffee}, \textit{Set Study Table}).



\begin{figure}[htpb]
\centering
\begin{subfigure}[t]{0.30\linewidth}
    \centering
    \includegraphics[width=\linewidth]{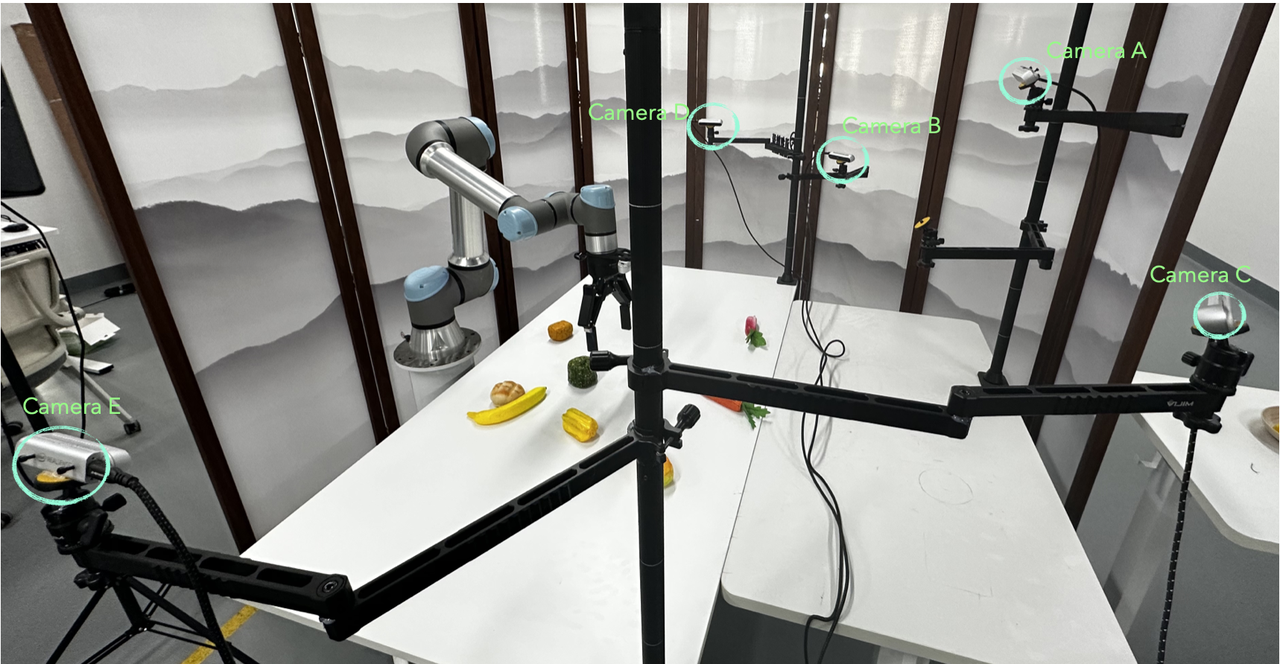}
    \caption{Real-world camera setup}
    \label{fig4a}
\end{subfigure}
\hfill
\begin{subfigure}[t]{0.60\linewidth}
    \centering
    \includegraphics[width=\linewidth]{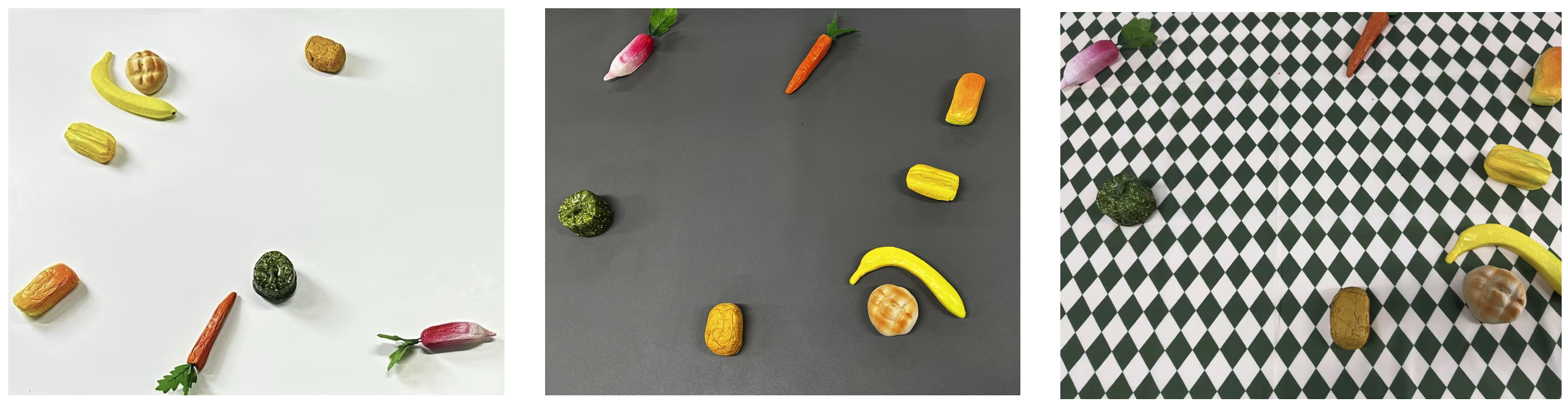}
    \caption{Backgrounds (ID, OOD-1, OOD-2)}
    \label{fig4b}
\end{subfigure}
\caption{\textbf{Real-World setting.} (a) Physical camera arrangement; (b) Different experimental backgrounds.}
\label{fig:combined_setup}
\vspace{-2mm}
\end{figure}

Real-world dataset contains 20,481 teleoperated demonstrations in 24 functionally diverse tasks as shown in \autoref{APPfig01} (5-15s duration distribution). We collect data through a calibrated UR5e robotic system with synchronized five camera sensing as shown in \autoref{fig4a}.





\vspace{-4mm}
\subsection{Training Configuration}

All models are trained with a consistent configuration to ensure fair comparison and reproducibility. 
For OmniD, DP,  VQ-BET\cite{lee2024behaviorgenerationlatentactions}, and ACT, the observation horizon is set to 16 steps, with 2 observation steps and 8 action steps per trajectory, capturing both short-term and long-term dependencies in sequential decision-making tasks. The batch size is 16, and the training step is 100k. The input image resolution is set to $480 \times 480$. To ensure reproducibility in all runs, we conducted all experiments using a fixed set of random seeds: \{1, 2, 3, 4\}. The optimizer used is AdamW, initialized with a learning rate of 1e-4. A cosine learning rate schedule with 500 warm-up steps is employed to facilitate smooth optimization dynamics during the early stages of training. 

\FloatBarrier
\section{Experiment}
In this section, we evaluate the methods' ID, OOD, and fine-tuning performance through both simulation and real-world experiments. In the ID experiment, training and evaluation are conducted in the same environment, with no changes to the background or object‑position distribution. In the OOD experiment, we alter both the object‑position distribution and the background.


\subsection{Simulation Experiment} 
\begin{figure}[htpb]
    \centering
    \begin{subfigure}[b]{\linewidth}
        \centering
        \includegraphics[width=\linewidth]{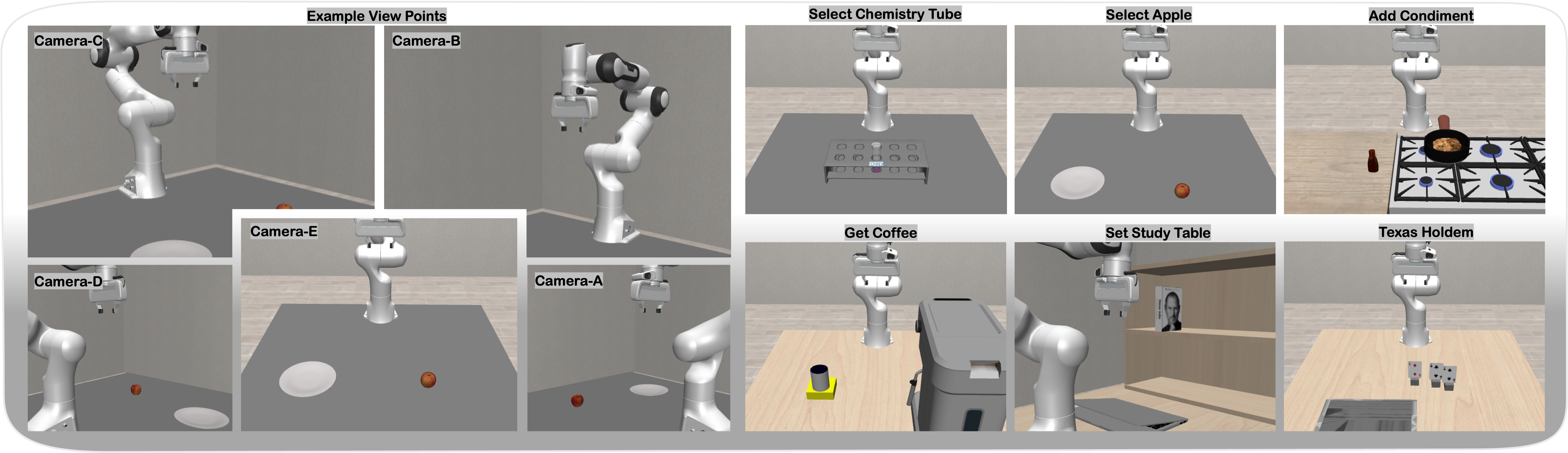}
        \vspace{-2mm}
        \caption{Visualization of the multiview setup alongside the six evaluation tasks in our benchmark.}
        \label{fig:train_eval_env_a}
    \end{subfigure}
    
    \vspace{1mm}  
    
    \begin{subfigure}[b]{\linewidth}
        \centering
        \includegraphics[width=\linewidth]{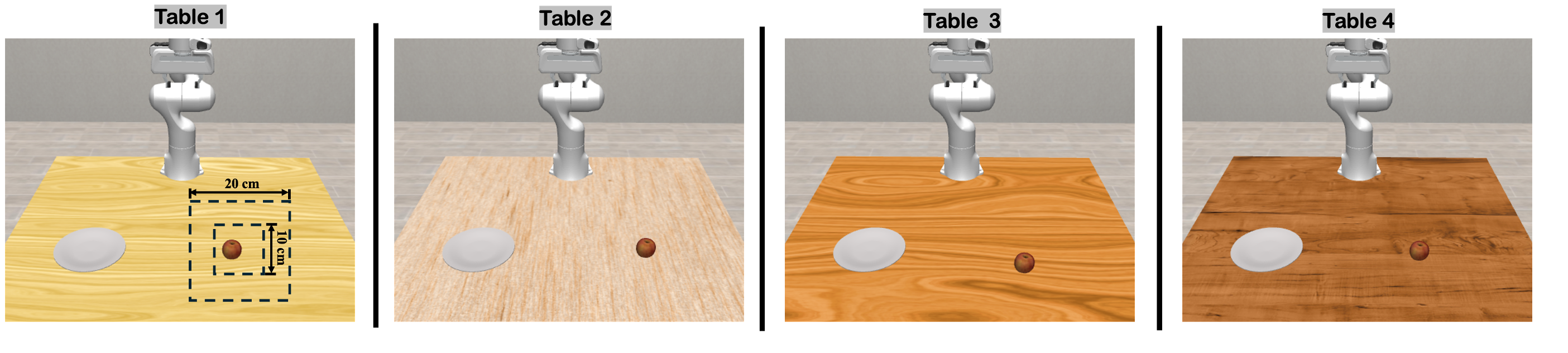}
        \vspace{-5mm}
        \caption{The background OOD setting and position OOD setting, in the table 1, the inner balck rectangle area is the distribution of object in traning set, the area between inner black retangle and outer one is the position distribution of object in evlation.}
        \label{fig:train_eval_env_b}
    \end{subfigure}
    
    \caption{Overview of our training and evaluation environments and task settings in VLAbench.}
    \label{fig:train_eval_env_combined}
    \vspace{-3mm}
\end{figure}

For our simulation experiments, we focus on six representative tasks: \textit{Select Apple}, \textit{Add Condiment}, \textit{Select Chemistry Tube}, \textit{Get Coffee}, \textit{Set Study Table}, and \textit{Texas Holdem} as shown on the right of \autoref{fig:train_eval_env_a}.

The bottom of \autoref{fig:train_eval_env_b} illustrates our OOD generalization setup. Background OOD is introduced by varying table textures, while object position OOD is created through procedural variation to simulate controlled distribution shifts. For position generalization, the black rectangle illustrates the spatial configuration: the inner 10 cm~$\times$~10 cm square represents the ID training region, while the surrounding annular band extending from 10 cm to 20 cm defines the OOD evaluation area. Note that the black rectangle is not a perspective-correct rendering and does not reflect the true distribution of the object positions. Instead, it serves as an approximate visual reference to indicate the spatial division between ID and OOD regions. These OOD environments do not overlap with the training set and are used exclusively for zero-shot evaluation.

For each task, we collect 50 expert trajectories under the ID setting. As shown on the left side of \autoref{fig:train_eval_env_a}, each task is captured from five fixed camera viewpoints, providing a diverse set of perspectives.

\label{sec:Simulation}

\setlength{\tabcolsep}{2pt}     

\newcolumntype{C}[1]{>{\centering\arraybackslash}p{#1}}
\begin{table*}[t]
\small
\renewcommand{\arraystretch}{1.2}
\centering
\begin{tabularx}{\textwidth}{l *{8}{>{\centering\arraybackslash}X} c}
\toprule
\textbf{Method}
  & \textbf{Select Apple}
  & \textbf{Add Condiment}
  & \textbf{Chemistry Tube}
  & \textbf{Get Coffee}
  & \textbf{Set Table}
  & \textbf{Texas Holdem} 
  & \textbf{Average} \\
\midrule
DP            & \bfseries 96 & \bfseries 100 & 70  &  88  &  48 & 90 & 82.0 \\ 
ACT           & 52           & 86            & 94  & 0            & 0            & 28 & 43.3 \\
VQ-BET        & 8           & 26            & 66  & 90            & 40            & 68 & 49.7  \\
$\pi$0 (LoRA)    & 12           & 72            & 92  & 12            & 0            & 72 & 43.3 \\
\midrule
\textbf{OmniD}
              & \bfseries 96 & \bfseries 100 & \bfseries 100
              & \bfseries 94 & \bfseries 56 & \bfseries 100 &\bfseries 91.0\\
\bottomrule
\end{tabularx}

\vspace{1mm}
\caption{
\textbf{Success rates (\%) on six benchmark tasks under multi-view (BCDE) input.}
OmniD leverages information from multiple cameras to achieve the highest performance across all tasks.
}
\label{tab:1}

\end{table*}
\subsubsection{In-distribution Performance}
We report simulation evaluation results in \autoref{tab:1}. All methods are trained and evaluated using multi-view inputs from cameras B, C, D, and E, which are collectively referred to as BCDE throughout the paper. OmniD consistently attains the highest success rates across all six task settings, demonstrating robust performance in both short‑ and long‑horizon manipulation tasks.
Compared to our baseline, DP, OmniD yields an average improvement of 9\% in success rate, with particularly notable gains on long-horizon and 3D manipulation tasks  such as \textit{Get Coffee} (from 88\% to 94\%) and \textit{Select Chemistry Tube} (from 70\% to 100\%). While DP performs competitively in tasks like \textit{Select Apple} and \textit{Add Condiment}, its performance drops in scenarios requiring extended spatial understanding.
ACT and VQ-BET serve as strong representatives of current state-of-the-art 2D behavior cloning methods. Although ACT achieves high success in \textit{Select Chemistry Tube} (94\%), it fails completely on \textit{Get Coffee} and \textit{Set Study Table}, highlighting its difficulty in generating consistent, multimodal action sequences over long horizons. Its tendency to produce abrupt or unrealistic motions suggests a lack of trajectory coherence in extended planning scenarios.
VQ-BET, while effective in encoding discrete visual-motor patterns, underperforms in spatially demanding 3D tasks ( \textit{Select Apple}: 8\%, \textit{Add Condiment}: 26\%), likely due to limited spatial accuracy and generalization capabilities.
OmniD outperforms all baselines across both 3D manipulation (\textit{Select Chemistry Tube}: 100\%) and long-horizon tasks (\textit{Get Coffee}: 94\%, \textit{Set Study Table}: 56\%, \textit{Texas Holdem}: 100\%). These gains are attributed to its efficient multi-view fusion in the form of implicit 3D feature representation and improved spatial reasoning, which together support robust policy generation even in tasks that require highly precise actions or involve significant view occlusion.
Lastly, despite having approximately 3.3 billion parameters, $\pi$0 (LoRA) exhibits inconsistent performance across tasks. While it performs reasonably well in certain scenarios, its overall performance remains unstable. Recent findings by \cite{liu2024rdt} have also reported the instability of large VLA models. We observe that $\pi$0 is highly sensitive to simulation stochasticity and often fails to converge reliably during fine-tuning, which likely contributes to its degraded performance in our evaluations.




\subsubsection{Out-of-distribution and fine-tuning performance }

We evaluated the models in \autoref{table_ood_sim} under three challenging OOD configurations and further conducted fine-tuning experiments to compare few-shot adaptation efficiency. All methods are first trained on task 0 (\textit{Select Apple}) using the BCDE viewpoints, followed by evaluation in out-of-distribution (OOD) scenarios involving spatial and background variations. Finally, the models trained on the BCDE viewpoints of task 0 are fine-tuned using 10 trajectories collected from the A viewpoint of task 0 and task 1 (\textit{Select Chemistry Tube}), to evaluate their capability for cross-scene adaptation

\textbf{Spatial generalization.} \autoref{fig:train_eval_env_b} illustrates the object position configurations used to train each policy
and the OOD object position configurations used during evaluation. As reported in \autoref{table_ood_sim}, OmniD achieved an 18\% success rate under OOD conditions. while baseline approaches achieved only 2\%, revealing a 9× performance gap in spatial generalization capability. Our OFG module is capable of learning robust 3D representations from multi-view input, which endows Omnid with enhanced spatial generalization capabilities.

\textbf{Background Generalization.}
We evaluated OmniD's robustness to background variations using four distinct OOD table textures, as illustrated in \autoref{fig:train_eval_env_b}. As reported in \autoref{table_ood_sim},  OmniD achieved a peak success rate of 90\% under the BCDE camera configuration, substantially outperforming baseline methods, whose best performance under the same setting was limited to 18\%.
This strong generalization is attributed to the deformable attention mechanism, which leverages learnable offsets to adaptively focus on object-centric regions. By avoiding distraction from highly variable background textures, this mechanism allows OmniD to extract more task-relevant spatial features, an essential capability for real-world deployment where environmental conditions may fluctuate significantly.

\textbf{Few-shot adaptability.} We conducted cross-view adaptation experiments using the models pretrained on camera configurations BCDE for \textit{Select Apple} task. The pretrained models were subsequently fine-tuned using 10 demonstration trajectories under two distinct settings: 1) intra-task cross-view adaptation (\textit{Select Apple} with camera configurations A), and 2) cross-task generalization (\textit{Select Chemistry Tube} with camera configurations A). Both configurations were rigorously evaluated through 50 trial episodes per task condition. As shown in \autoref{table_ood_sim}, OmniD achieved success rates of 88\% and 80\%, respectively, while the baseline models failed to adapt to the novel perspective and task, with the best-performing one reaching only a 2\% success rate.


\begin{table}[htpb]
\centering
\small
\renewcommand{\arraystretch}{1.2}
\begin{tabularx}{\textwidth}{l *{5}{>{\centering\arraybackslash}X}}
\toprule

\textbf{Method}
& \makecell{\textbf{Task 0}\\\textbf{ID}}
& \makecell{\textbf{Position}\\\textbf{OOD}}
& \makecell{\textbf{Background}\\\textbf{OOD}}
& \makecell{\textbf{Fine-tune}\\\textbf{Task 0-A}}
& \makecell{\textbf{Fine-tune}\\\textbf{Task 1-A}} \\
\midrule
DP            & 96                      & 0                    & 0/0/0/0                 & 0                              & 0                              \\
ACT           & 52                      & 0                    & 0/0/0/0                 & 2                              & 0                              \\
VQ-BET        & 4                       & 0                    & 4/6/0/4                 & 0                              & 0                              \\
$\pi$0 (LoRA)    & 28                      & 2                    & 6/18/12/8               & 0                              & 0                              \\
\midrule
\textbf{OmniD}        & \textbf{96}             & \textbf{18}          & \textbf{0/90/82/20}     & \textbf{88}                   & \textbf{80}                   \\
\bottomrule
\end{tabularx}
\vspace{4mm}
\caption{\textbf{Performance comparison under OOD conditions and fine-tuning.} Task 0 and Task 1 represent \textit{Select Apple} and \textit{Select Chemistry Tube} respectively. Position OOD evaluates on task 1  with randomized spatial placements in novel configurations, while background OOD assesses performance with four distinct desktop backgrounds. Success rates (\%) are reported for all methods.}
\vspace{-3mm}
\label{table_ood_sim}
\end{table}

\normalsize






\vspace{-3mm} 
\subsection{Real-world Task}
We conduct comparative real-world evaluations with the baseline DP on the \textit{pickup pumpkin} task. Our method exhibits highly consistent performance across both ID and OOD evaluations as well as in simulation, substantially outperforming all baseline approaches. The experiment setting is shown in \autoref{fig:mainfig}. We collect 50 episodes for pretraining and use the same hyperparameters and training steps as in the simulation experiments. 

As show in \autoref{tab3}, OmniD consistently outperforms the baseline DP across diverse scenarios and perspectives.
All models are evaluated 25 times, and then calculate the average success rate.
As shown in \autoref{tab3}, our model achieves a significantly higher success rate under the BCDE camera setup, reaching 84\%, whereas the baseline model achieves just 24\%.
\begin{table}[htb]
\centering
\renewcommand{\arraystretch}{1.2} 
\begin{tabularx}{\textwidth}{l *{6}{>{\centering\arraybackslash}X}}
        \toprule
\textbf{Method} & \textbf{Training} & \textbf{Evaluation} & \textbf{Original} & \textbf{OOD-1} & \textbf{OOD-2} &\textbf{Fine-tune on A} \\ \hline
DP    & BCDE & BCDE & 24 & 0  & 0& 0 \\

\hline
\textbf{OmniD} & BCDE & BCDE & \textbf{84} & \textbf{24} & \textbf{8} & \textbf{60} \\
\bottomrule
\end{tabularx}
\vspace{4mm}
\caption{\textbf{realworld performance.} We evaluate the model's success rate on different backgrounds. In the last column, we present the success rate after fine-tuning on 10 trajectories from a new perspective.}
\label{tab3}
\end{table}
\normalsize



\FloatBarrier
\vspace{-6mm}
We have three types of background, as shown in \autoref{fig4b}. 
The experimental results in \autoref{tab3} demonstrate the generalizability of OmniD across varying backgrounds, compared to the baseline method. OmniD maintains moderate accuracy on OOD-1 (24\%) and OOD-2 (8\%) backgrounds, whereas DP fails to generalize, scoring 0\% on both.

The fine-tuning results in \autoref{tab3} further underscore the adaptability of OmniD compared to DP. When fine-tuned on the A setup with only 10 trajectories after training on BCDE, OmniD achieves a notable performance of 60\% on the original background, while DP fails with 0\%. This demonstrates OmniD’s ability to leverage fine-tuning to improve generalization across different setups, highlighting its data efficiency and adaptability to real-world scenarios.

\subsection{Ablation Analysis}
\begin{table}[htbp]
  \centering

  \begin{tabularx}{\textwidth}{l *{4}{>{\centering\arraybackslash}X}}
    \toprule
     & \textbf{ResNet-101} & \textbf{w/o OFG} & \textbf{Camera A} & \textbf{Base Model} \\
    \midrule
    ResNet-18      &  & \checkmark             &    \checkmark          &    \checkmark           \\
    ResNet-101      & \checkmark &            &            &             \\
    OFG module                & \checkmark &            & \checkmark    & \checkmark  \\
    Multi-view input fusion   & \checkmark & \checkmark &            & \checkmark  \\
    \midrule
    \textbf{Success Rate (\%)}& 97.3       & 84.3       & 84.0       & \textbf{96.0} \\
    \bottomrule
  \end{tabularx}
  \vspace{5mm}
    \caption{Ablation study evaluating the effects of backbone choice, the OFG module, and camera configuration on task success rates. ResNet-101 achieves the highest accuracy (97.3 \%), while our base model (ResNet-18 with OFG and BCDE multi-view fusion) attains 96.0 \%.}
  \vspace{-2mm}
  \label{tab:ablation}
\end{table}
\vspace{-2mm}
\normalsize
To investigate the effectiveness of each architectural component and the multi-view input design, we conduct an ablation study summarized in Table~\ref{tab:ablation}. Using ResNet-18 as the backbone, we evaluate the contribution of the OFG module and the multi-view input. Removing OFG results in a notable drop in performance from 96.0\% to 84.3\%, highlighting its role in enhancing spatial reasoning around task-relevant objects.
We also assess the impact of multi-view observations by comparing single-view input (Camera A) against multi-view input (Cameras BCDE). The single‐view configuration attains only 84.0 \%, whereas integrating multiple views restores performance to 96.0 \%, underscoring the efficacy of multi‐view feature fusion. Additionally, we compare ResNet-18 with a deeper ResNet-101 backbone, which yields a marginal improvement to 97.3\%, suggesting that our architecture is not heavily dependent on backbone capacity.

\vspace{-3mm} 
\section{Conclusion} 
\vspace{-3mm} 
\label{sec:conclusion}
In this paper, we present OmniD, a novel framework designed to address the critical challenges of poor generalization capabilities and inefficient fine-tuning processes in current 2D visuomotor methods. Without relying on costly point cloud or depth information, OmniD implicitly extracts 3D features from multi-view inputs by leveraging a deformable attention mechanism. This approach allows the model to focus more on the target objects and less on the irrelevant background.  The 3D feature significantly improves generalization performance, making it more adaptable across varying environments and camera setups. Our experiments demonstrated the efficacy of OmniD in our simulation and real-world benchmark. In the simulation benchmark, OmniD achieves
9\%, 33.8\%, and 83\% average improvement over the best baseline model for ID, OOD, and few-shot experiments, respectively.
In the real-world experiment, OmniD outperformed the best baseline model in multi-view fusion with an 84\% success rate compared to 24\%. In background generalization, OmniD achieved a success rate of 24\%, while the base models completely failed. For new camera fine-tuning, OmniD achieved 60\% success with only 10 samples for fine-tuning, while the baseline got 0. Our work provides strong evidence for the effectiveness of constructing 3D representations from image inputs in robotic manipulation tasks

\section{Limitations} 
Our framework has two main limitations: \textbf{Dependence on Camera Calibration}: Although it does not rely on depth information, our method still requires approximate camera extrinsics when constructing 3D features. Moderate calibration errors can be tolerated without significant degradation in performance, even in dynamic environments. \textbf{Limited Testing on Multi-Task Scenarios}: While we have demonstrated the efficiency and task generalization capabilities of OmniD through training on Task A and fine-tuning on Task B, our experiments have primarily focused on single-task settings. Further testing is required to evaluate the model's ability to generalize and perform across multiple tasks simultaneously, particularly in more complex dual-arm, long-horizon tasks. \textbf{Lack of exploration of the CD}: Exploring the generalizability of different models in the CD scenarios is a complex task, and large-scale training on diverse tasks and scenarios is required to achieve such generalization ability. As a work to explore the effectiveness of 3D BEV features in manipulation tasks, the experiments are designed to evaluate the models' performance in ID and OOD scenarios.






\bibliography{ref}  
\end{document}